\documentclass[11pt,a4paper]{article}

\usepackage[utf8]{inputenc}
\usepackage[T1]{fontenc}
\usepackage{lmodern}
\usepackage[margin=1in]{geometry}
\usepackage{amsmath,amssymb}
\usepackage{booktabs}
\usepackage{tabularx}
\usepackage{graphicx}
\usepackage{tikz}
\usepackage{pgfplots}
\pgfplotsset{compat=1.18}
\usepackage{hyperref}
\usepackage{natbib}
\usepackage{enumitem}
\usepackage{xcolor}
\usepackage{microtype}
\usepackage{float}

\hypersetup{
  colorlinks=true,
  linkcolor=blue!60!black,
  citecolor=blue!60!black,
  urlcolor=blue!60!black,
}

\title{\textbf{Umwelt Engineering: Designing the Cognitive Worlds\\of Linguistic Agents}}
\author{Rodney Jehu-Appiah}
\date{}

\begin{document}
\maketitle

\begin{abstract}
A tick's world contains butyric acid, temperature, and tactile density---not because other features do not exist, but because nothing else exists in its world. Jakob von Uexk\"{u}ll called this the organism's \emph{Umwelt}: the perceptual world its biology makes available. A language model reasons \emph{in} language. A human thinks across many modalities---spatial intuition, emotional valence, muscle memory, mental imagery---and uses language as one channel among several to articulate the result. A standard LLM has no such multiplicity. Its cognition unfolds entirely in the token stream: the words do not describe the thinking; they are the thinking. Change the available language and you change the cognition itself. Yet the field treats this medium as transparent, optimizing what agents are asked (prompt engineering) and what they know (context engineering) while leaving the linguistic world they think in unexamined.

I propose \emph{Umwelt engineering}---the deliberate design of the linguistic cognitive environment---as a third layer in the agent design stack, upstream of both prompt and context. Two experiments test the thesis that altering the medium of reasoning alters cognition itself. In Experiment~1, three language models (Claude Haiku~4.5, GPT-4o-mini, Gemini~2.5 Flash Lite) reason under two vocabulary constraints---No-Have (eliminating possessive ``to have'') and E-Prime (eliminating ``to be'')---across seven tasks ($N{=}4{,}470$ trials). The constraints do not uniformly help or hurt, but the pattern is striking. No-Have---which removes possessive framing from the agent's available language---improves ethical reasoning by 19.1~pp ($p < 0.001$), classification by 6.5~pp ($p < 0.001$), and epistemic calibration by 7.4~pp, while achieving 92.8\% constraint compliance and producing consistent benefits across models. E-Prime shows a more volatile profile: dramatic gains on causal reasoning (+14.1~pp) and ethical dilemmas (+15.5~pp), but model-dependent effects so severe that the same constraint improves Gemini's ethical reasoning by 42.3~pp while collapsing GPT-4o-mini's epistemic calibration by 27.5~pp. Cross-model correlations of E-Prime effects reach $r = -0.75$---evidence that different models occupy different native Umwelten shaped by their training, and that an imposed constraint interacts with each model's native world rather than overriding it. In Experiment~2, 16 agents, each constrained to a distinct linguistic mode, tackle 17 debugging problems at temperature~0.0 (deterministic). No constrained agent outperforms the control individually, yet a 3-agent ensemble selected for linguistic diversity achieves 100\% ground-truth coverage versus 88.2\% for the control---a result that depends critically on the counterfactual agent, which is the only agent to surface the hardest finding. A permutation test confirms: only 8\% of random 3-agent subsets achieve full coverage, and every one of them includes the counterfactual agent.

Two mechanisms emerge: \emph{cognitive restructuring}, where removing linguistic defaults forces more explicit reasoning---No-Have's removal of possessive framing produces the broadest and most consistent restructuring, while E-Prime's removal of the copula produces deeper but less predictable effects---and \emph{cognitive diversification}, where different constraints activate different regions of a model's latent capacity. Together, they establish the linguistic medium of agent reasoning as a first-class design variable and open a structured research agenda for the systematic construction of cognitive environments for artificial minds. The primary methodological limitation is the absence of an active control matching the constraint prompts' elaborateness without imposing a vocabulary restriction; the crossover pattern (task-specific improvements and degradations) is inconsistent with a generic instruction effect, but cannot fully rule out a metalinguistic self-monitoring confound.
\end{abstract}

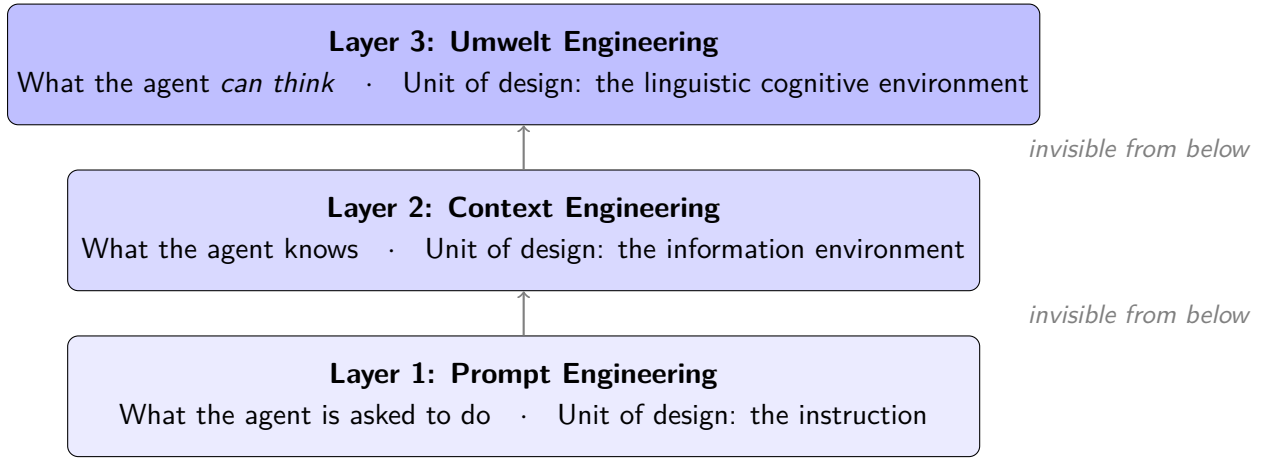
\begin{figure}[t]
\centering
\begin{tikzpicture}[
  layer/.style={draw, rounded corners=4pt, minimum width=12cm, minimum height=1.6cm, align=center, font=\sffamily},
]
\node[layer, fill=blue!8] (L1) at (0,0) {\textbf{Layer 1: Prompt Engineering}\\[2pt]What the agent is asked to do\quad$\cdot$\quad Unit of design: the instruction};
\node[layer, fill=blue!15] (L2) at (0,2.2) {\textbf{Layer 2: Context Engineering}\\[2pt]What the agent knows\quad$\cdot$\quad Unit of design: the information environment};
\node[layer, fill=blue!25] (L3) at (0,4.4) {\textbf{Layer 3: Umwelt Engineering}\\[2pt]What the agent \emph{can think}\quad$\cdot$\quad Unit of design: the linguistic cognitive environment};

\draw[->, thick, gray] (L1.north) -- (L2.south);
\draw[->, thick, gray] (L2.north) -- (L3.south);

\node[font=\small\sffamily\itshape, text=gray, anchor=west] at (6.5,1.1) {invisible from below};
\node[font=\small\sffamily\itshape, text=gray, anchor=west] at (6.5,3.3) {invisible from below};
\end{tikzpicture}
\caption{The three-layer stack for AI agent design. Each layer is invisible from the one below it: a prompt engineer does not reason about the linguistic structures through which prompts are interpreted. Umwelt engineering operates on the vocabulary, grammar, and conceptual primitives that constitute the agent's cognitive world.}
\label{fig:stack}
\end{figure}

\section{Introduction}

The practice of building effective AI agents has consolidated around two disciplines. \emph{Prompt engineering} optimizes the formulation of requests to elicit desired outputs. \emph{Context engineering} manages the information environment at inference time---retrieval-augmented generation, tool results, memory systems, and system prompts. Both disciplines treat language as transparent---a vehicle for carrying instructions and information, not itself a variable that determines what the agent can think.

This assumption is untenable. Mounting evidence demonstrates that linguistic structure---independent of informational content---systematically alters the reasoning behavior of large language models (LLMs). Chinese-trained and English-trained models internalize different causal reasoning patterns, not merely different surface forms \citep{wang2025}. Prompt formatting alone (plain text vs.\ Markdown vs.\ JSON) produces performance swings of up to 40\% on identical tasks \citep{he2024}. Models trained with internal ``thought tokens'' develop emergent reasoning formats that outperform natural language chain-of-thought \citep{zelikman2024}. And synthetic reasoning languages designed for LLM cognition achieve 4--16$\times$ token compression with near-parity accuracy \citep{tanmay2025}.

These findings converge on a single insight: for a language model, the available language is not a medium through which cognition passes---it \emph{is} the cognition. I propose the term \emph{Umwelt engineering} for the deliberate design of this linguistic substrate, and argue that it constitutes a third layer in the agent design stack, upstream of both prompt and context engineering (Figure~\ref{fig:stack}).

\subsection{The Umwelt Concept}

Jakob von Uexk\"{u}ll \citeyearpar{uexkull1934} introduced the \emph{Umwelt} to describe the perceptual world of an organism---not the objective environment, but the subset of reality that the organism's biology makes available to it. A tick's Umwelt contains butyric acid, temperature, and tactile hair density; a bat's Umwelt is structured by echolocation returns. Each organism inhabits a different world, not because the physical environment differs, but because its sensory apparatus admits different features.

For a biological organism, the Umwelt is a \emph{filter}---a subset of a richer physical reality, selected by biology. A human thinks across many modalities---visual imagery, proprioception, emotion, spatial reasoning---and often reaches a conclusion before finding words for it. Language is one cognitive channel among several. For a standard large language model, the relationship between language and cognition is not filtering but identity. The model's reasoning unfolds entirely in the token stream. The words do not report on cognition that happened elsewhere; they are the cognition. An LLM's Umwelt, therefore, is not a filtered view of some deeper cognitive space---it is the cognitive space in its entirety.

I define an LLM's Umwelt as the totality of the linguistic structures available to it at inference time: the vocabulary it can deploy, the grammatical patterns it can instantiate, the conceptual distinctions those patterns make expressible. Change these, and you do not filter the agent's perception of its thoughts---you change what thoughts it can have.

This makes the Umwelt concept apply more completely to language models than to the biological organisms for which it was coined. A tick has a body operating below its perceptual world---metabolic processes, locomotion, behaviors that its Umwelt does not represent. A human has spatial reasoning, proprioception, affect---entire cognitive systems that operate without language. A standard LLM has no such sub-linguistic remainder. The language goes all the way down. When you remove ``to be'' from an agent's available vocabulary, you do not ask it to ignore a perceptual channel it still possesses---you remove a class of cognitive operations from the only substrate in which its cognition occurs.

Critically, the boundary of the Umwelt is not a barrier the agent strains against---it is invisible. A tick does not experience the absence of color vision. An agent reasoning without the concept of ``epistemic tension'' does not notice when two of its beliefs conflict; the conflict is not suppressed but absent as a category of perception. This is what distinguishes Umwelt engineering from prompting: you are not instructing the agent to think differently, you are constituting the world in which it thinks.

\subsection{The Three-Layer Stack}

I propose the following abstraction hierarchy for AI agent design:

\begin{enumerate}[nosep]
  \item \textbf{Prompt engineering}---optimizing what the agent is asked to do. Unit of design: the instruction.
  \item \textbf{Context engineering}---optimizing what the agent knows at inference time. Unit of design: the information environment.
  \item \textbf{Umwelt engineering}---optimizing what the agent \emph{can think}. Unit of design: the linguistic cognitive environment.
\end{enumerate}

Each layer is invisible from the one below it. A prompt engineer does not reason about memory architectures. A context engineer does not reason about whether the agent should possess the concept of a counterfactual. And an Umwelt engineer designs the linguistic world---the vocabulary, grammar, conceptual primitives, and reasoning structures---within which all prompts and all context are interpreted.

\section{Related Work}

\subsection{Linguistic Relativity in LLMs}

The Sapir-Whorf hypothesis---that language shapes thought---has been empirically tested in LLMs with affirmative results. \citet{wang2025} created BICAUSE, a bilingual causal reasoning dataset, and demonstrated that LLMs internalize language-specific reasoning biases: Chinese-trained models focus attention on causes and sentence-initial connectives, while English-trained models show balanced distributions. Models rigidly apply these patterns even to atypical inputs, degrading performance when task structure mismatches training language structure. \citet{ray2025} confirmed linguistic relativity effects in GPT-4o across culturally salient prompts. \citet{alkhamissi2025} tracked 34 training checkpoints and found that while early training aligns LLMs with human language processing, advanced models diverge---suggesting they develop their own cognitive relationship to language rather than merely mimicking human patterns.

\subsection{Designed Reasoning Languages}

ORION \citep{tanmay2025} explicitly implements Fodor's Language of Thought Hypothesis for LLMs, creating ``Mentalese''---a symbolic reasoning format where each step is serialized as \texttt{OPERATION:expression;} (SET, CALC, EQ, SOLVE, ANS). Using the MentaleseR-40k dataset, models reasoning in Mentalese achieve 4--16$\times$ fewer tokens and up to 5$\times$ lower inference latency with 90--98\% accuracy retention. This constitutes a direct demonstration that designing the reasoning language alters cognitive performance.

Sketch-of-Thought \citep{sketch2025} introduces three cognitively-inspired reasoning paradigms---Conceptual Chaining, Chunked Symbolism, and Expert Lexicons---with a routing model selecting the appropriate paradigm per task. Token reductions reach 84\% with maintained or improved accuracy, demonstrating that different tasks benefit from different cognitive dialects.

\subsection{Beyond Linguistic Reasoning}

Coconut \citep{hao2024} removes language from reasoning entirely by feeding hidden states back as input embeddings, enabling breadth-first exploration of reasoning paths rather than the linear commitment enforced by token-by-token generation. Quiet-STaR \citep{zelikman2024} trains models to generate internal ``thought tokens'' before each output token, developing an emergent reasoning format that doubled math performance on GSM8K. Both approaches suggest that natural language may constrain as much as it enables---a finding consistent with the Umwelt framework, which predicts that what an agent can think is determined by what it can think \emph{in}.

\subsection{Cognitive Linguistics and AI}

\citet{kramer2025} applied Lakoff and Johnson's Conceptual Metaphor Theory (CMT) as a prompting paradigm, using metaphorical source-domain mappings to structure abstract reasoning. CMT-augmented models significantly outperformed baselines across domain-specific reasoning, creative insight, and metaphor interpretation tasks. This establishes that cognitive-linguistic structures---not just informational content---serve as effective reasoning affordances for LLMs.

\subsection{E-Prime and General Semantics}

E-Prime, developed by David Bourland Jr.\ \citeyearpar{bourland1965} as an application of Alfred Korzybski's \citeyearpar{korzybski1933} general semantics, eliminates all forms of ``to be'' from English. The theoretical motivation is that the copula enables identity-level assertions (``X \emph{is} Y'') that conflate map and territory---treating descriptions as essences. E-Prime forces operational reformulation: ``this code is buggy'' becomes ``this code produces incorrect output when given input X.'' While studied in human communication and pedagogy \citep{bourland1991}, E-Prime has not been tested as a reasoning constraint for LLMs prior to this work.

\subsection{A Taxonomy of Linguistic Constraint Traditions}

E-Prime is one instance of a broader phenomenon: intellectual traditions that identified specific axes along which language shapes thought, and proposed linguistic reforms to intervene. I survey eight such traditions, each targeting a distinct cognitive axis, to establish the theoretical foundation for a principled constraint design space.

\textbf{General Semantics} \citep{korzybski1933}. Beyond E-Prime, Korzybski's full system includes extensional devices: indexing (Smith$_1 \neq$ Smith$_2$---no two referents of the same word are identical), dating (the economy$_{2024} \neq$ the economy$_{2026}$---referents change over time), and the structural differential (every description omits detail---append ``etc.''\ to maintain map-territory awareness). These devices target \emph{over-generalization}: the tendency to treat a label as if it captured the full structure of its referent.

\textbf{Rheomode} \citep{bohm1980}. The physicist David Bohm proposed a mode of English in which verbs are primary and nouns are derived. Standard English reifies process into entity: ``the electron moves'' presupposes a static thing that then acts. Bohm's rheomode reverses this, treating movement as fundamental and the electron as an abstraction drawn from it. For LLM reasoning, a simplified rheomode constraint---``express everything as process; no static noun-based assertions''---targets \emph{entity bias}: the tendency to reason about systems as collections of fixed objects rather than ongoing transformations.

\textbf{Operationalism} \citep{bridgman1927}. The physicist Percy Bridgman argued that every scientific concept must be defined by the operations used to measure it. ``Length'' means the result of applying a measuring rod; ``simultaneity'' means the outcome of a specific synchronization procedure. A concept without an operational definition is, for Bridgman, meaningless. Applied as a linguistic constraint, operationalism targets \emph{ungrounded claims}---assertions that sound precise but lack any connection to observable procedure.

\textbf{Constructed Languages for Cognitive Intervention.} Two constructed languages directly implement Whorfian interventions. \emph{Lojban} \citep{brown1955}, derived from predicate logic, eliminates syntactic ambiguity entirely---every sentence has exactly one parse, forcing the speaker to commit to precise logical structure. \emph{Toki Pona} \citep{lang2001}, with approximately 130 words, forces radical decomposition of complex concepts into primitive components. A ``computer'' becomes a ``knowledge tool''; a ``hospital'' becomes a ``body-fixing house.'' As a reasoning constraint, Toki Pona targets \emph{abstraction leakage}: the tendency to hide incomplete understanding behind technical vocabulary.

\textbf{Grammatical Evidentiality} \citep{elgin1984}. Suzette Haden Elgin's constructed language L\'{a}adan, created for her novel \emph{Native Tongue}, includes obligatory evidentiality markers: every statement must be grammatically tagged with how the speaker knows it---direct observation, inference, hearsay, assumption, or dream. Applied as a constraint on LLM reasoning, mandatory evidentiality tagging targets \emph{epistemic opacity}---the tendency of models to present inferences, training priors, and confabulations in the same assertive voice as direct textual evidence.

\textbf{Catu\d{s}ko\d{t}i} \citep{nagarjuna150}. The Buddhist tetralemma admits four truth values for any proposition: true, false, both true and false, and neither true nor false. N\={a}g\={a}rjuna's \emph{M\={u}lamadhyamak\={a}rik\={a}} uses this four-valued logic to examine and reject essentialist claims about causation, identity, and existence. As a reasoning constraint, the catu\d{s}ko\d{t}i targets \emph{premature binary resolution}---the tendency to collapse complex or paradoxical situations into yes/no answers when the evidence supports a more nuanced position.

\textbf{Nonviolent Communication} \citep{rosenberg2003}. Marshall Rosenberg's NVC framework structures all communication into four components: observation (what happened, without evaluation), feeling (the speaker's emotional response), need (the underlying value at stake), and request (a concrete action). The critical discipline is the first step: separating observation from judgment. As a reasoning constraint, NVC targets \emph{conflation of observation with evaluation}---a failure mode particularly relevant to code review, architectural critique, and any task where premature judgment short-circuits analysis.

These eight traditions span at least seven distinct axes of linguistic intervention: identity claims (E-Prime), over-generalization (General Semantics), entity bias (Rheomode), ungrounded abstraction (Operationalism), syntactic ambiguity (Lojban), lexical compression (Toki Pona), epistemic sourcing (L\'{a}adan), binary logic (Catu\d{s}ko\d{t}i), and observation-judgment conflation (NVC). The existence of this many independently motivated traditions, each targeting a different cognitive failure mode through linguistic reform, constitutes prima facie evidence that the design space of cognitive-linguistic constraints is rich, structured, and largely unexplored in the context of artificial agents.

\textbf{A note on methodology.} The intuition behind this paper has two sources. The first is personal: as a speaker of both English and Kasem, a Gur language of northern Ghana whose grammatical structures differ substantially from English---in how it encodes time, causation, and social obligation---I experienced linguistic relativity not as an academic hypothesis but as a fact of cognition. Reasoning about the same problem in different languages produced different reasoning, not just different words. The second is a prior interest in E-Prime, which suggested that this effect could be \emph{engineered} within a single language by selectively removing grammatical structures. The hypothesis that a broader, principled design space of such interventions existed motivated a directed literature survey using Claude (Anthropic, 2024--2026) as a research tool, much as one might use a domain expert to identify candidate traditions across fields one has not studied directly. Claude surfaced the specific traditions; I evaluated each for relevance, mapped it to a constraint axis, and integrated it into the framework.

\section{Experiment 1: Linguistic Constraints as Cognitive Interventions}

\subsection{Design}

I test the hypothesis that constraining an LLM's reasoning language alters performance in task-dependent ways. Two linguistic constraints are tested, each targeting a different axis of linguistic default.

\textbf{E-Prime} eliminates all forms of ``to be'' (is, am, are, was, were, be, being, been, and contractions: it's, that's, there's, who's). This removes identity assertions as a grammatical possibility, forcing the model to reformulate reasoning in operational, behavioral, or relational terms. Theoretical source: \citet{korzybski1933}, \citet{bourland1965}.

\textbf{No-Have} eliminates all forms of ``to have'' used as a main verb (has, have, had, having), excluding auxiliary uses (e.g., ``has completed'' remains permitted). This removes possessive framing, forcing the model to replace ownership/containment language with relational, behavioral, or structural descriptions. Theoretical source: the broader General Semantics program of identifying linguistic defaults that encode cognitive defaults.

\textbf{Predictions.} No-Have was predicted to improve tasks saturated with possessive framing---ethical dilemmas (patients ``have'' rights, actions ``have'' consequences), classification (categories ``have'' members), and epistemic calibration (claims ``have'' support)---and show neutral effects on tasks where possession is incidental to the reasoning structure. E-Prime was predicted to \emph{degrade} performance on tasks with native ontological structure (syllogisms run on ``X is a Y''---circumlocution adds friction) and \emph{improve} performance on tasks where essentialist shorthand masks reasoning gaps (causal reasoning, where ``the cause is X'' collapses mechanism articulation; ethical dilemmas, where ``X is wrong'' forecloses analysis). Math word problems were predicted to show no effect under either constraint (numerical reasoning is largely independent of copula or possessive structure).

\textbf{Models.} Three models from three providers: Claude Haiku~4.5 (\texttt{claude-haiku-4-5-20251001}, Anthropic), GPT-4o-mini (\texttt{gpt-4o-mini-2024-07-18}, OpenAI), and Gemini~2.5 Flash Lite (\texttt{gemini-2.5-flash-lite}, Google). All are cost-efficient instruction-following models of comparable capability. Cross-vendor replication tests whether constraint effects are properties of linguistic structure or artifacts of a single model's training.

\textbf{Conditions.} (1)~Control: standard English, no constraints. (2)~E-Prime: explicit prohibition of all ``to be'' forms with grammatical enforcement instructions. (3)~No-Have: explicit prohibition of possessive ``to have'' forms.

\textbf{Tasks.} Seven task types (130 items total): syllogistic reasoning~(20), causal reasoning~(15), analogical reasoning~(20), classification~(20), epistemic calibration~(20), ethical dilemmas~(15), and math word problems~(20). All tasks use A/B/C/D multiple-choice format for scoring consistency (syllogisms use VALID/INVALID). Items span three difficulty levels.

\textbf{Procedure.} Each item was administered under each condition, for each model, with 4 repetitions: 1 at temperature~0.0 (deterministic) and 3 at temperature~0.7 (stochastic). Total design: $130 \times 3 \times 3 \times 4 = 4{,}680$ planned trials. Maximum 2,048 output tokens per trial. The experiment ran autonomously over approximately 5.5~hours with results flushed to disk after every trial.

\textbf{Metrics.} Binary accuracy (correct/incorrect per item), linguistic compliance (constraint violations per trial), word count, reasoning chain depth (step marker count), and epistemic specificity (ratio of grounded assertions to bare assertions).

\subsection{Results}

Of 4,680 planned trials, 4,470 completed successfully (30 Gemini 503 errors and 180 timeout/rate-limit failures). Of completed trials, 4,429 produced parseable responses, and 4,344 yielded extractable answers for accuracy scoring. The remaining 85 extraction failures were distributed across conditions without strong systematic bias (see Section~\ref{sec:exp1-limitations}).

\subsubsection{Aggregate Accuracy}

\begin{table}[t]
\centering
\caption{Aggregate accuracy by task and condition (all models pooled, $N{=}4{,}344$ scoreable trials). pp = percentage points. $p$-values from Fisher's exact test (two-sided). Effect sizes (Cohen's $d$, approximate for binary outcomes): ethical dilemmas $d = 0.57$ (No-Have), $d = 0.44$ (E-Prime); causal reasoning $d = 0.39$ (E-Prime); classification $d = 0.36$ (No-Have). Bootstrap 95\% CIs for ethical dilemmas No-Have delta: $[+12.1\%, +26.2\%]$; causal reasoning E-Prime delta: $[+5.6\%, +23.2\%]$.}
\label{tab:aggregate}
\smallskip
\begin{tabular}{@{}lcrcrcrc@{}}
\toprule
Task & Control & No-Have & $\Delta$(NH) & $p$(NH) & E-Prime & $\Delta$(EP) & $p$(EP) \\
\midrule
Ethical dilemmas      & 76.6\% & 95.6\% & +19.1 pp & ${<}0.001$*** & 92.1\% & +15.5 pp & ${<}0.001$*** \\
Classification        & 93.0\% & 99.6\% & +6.5 pp  & ${<}0.001$*** & 96.2\% & +3.1 pp  & ns \\
Epistemic calibration & 68.7\% & 76.1\% & +7.4 pp  & ns            & 63.0\% & $-$5.7 pp & ns \\
Causal reasoning      & 76.7\% & 81.5\% & +4.9 pp  & ns            & 90.8\% & +14.1 pp & ${<}0.001$*** \\
Math word problems    & 92.5\% & 93.9\% & +1.5 pp  & ns            & 92.8\% & +0.4 pp  & ns \\
Analogical reasoning  & 76.2\% & 74.9\% & $-$1.4 pp & ns           & 73.2\% & $-$3.0 pp & ns \\
Syllogisms            & 100.0\%& 97.9\% & $-$2.1 pp & 0.074        & 96.6\% & $-$3.4 pp & 0.015* \\
\bottomrule
\end{tabular}
\end{table}

Overall accuracy: Control 83.5\%, No-Have 88.6\%, E-Prime 85.4\%. No-Have produced the larger and more consistent improvement, raising accuracy on 5 of 7 tasks while achieving 92.8\% constraint compliance (compared to E-Prime's 48.1\%). E-Prime's high violation rate (51.9\%) means its effects reflect a mixture of compliant and non-compliant reasoning.

The most striking result belongs to No-Have. Removing possessive framing improved ethical dilemmas by 19.1~pp ($p < 0.001$), classification by 6.5~pp ($p < 0.001$), and epistemic calibration by 7.4~pp---a broad pattern of improvement that held across models with minimal degradation elsewhere (only analogical reasoning and syllogisms showed small, non-significant declines). No-Have outperformed E-Prime on 5 of 7 tasks, a result that was not predicted: No-Have was originally included as an exploratory second constraint.

E-Prime showed a more volatile profile. Its strongest single effect---causal reasoning +14.1~pp ($p < 0.001$)---exceeded No-Have on that task, and it improved ethical dilemmas by 15.5~pp ($p < 0.001$). But E-Prime degraded syllogisms ($-3.4$~pp, $p = 0.015$) and epistemic calibration ($-5.7$~pp), consistent with the prediction that removing ``to be'' impairs tasks whose structure depends on identity bridges and calibrated hedging. The crossover pattern---improvement on some tasks, degradation on others---is a signature of cognitive restructuring rather than generic facilitation or generic impairment.

Ethical dilemmas showed the largest effect under either constraint, suggesting that both identity framing (``X is wrong'') and possessive framing (``the patient has a right to\ldots'') actively impede ethical reasoning. The No-Have effect (+19.1~pp) exceeded E-Prime (+15.5~pp) on this task, suggesting possessive reification may be the more distorting default.

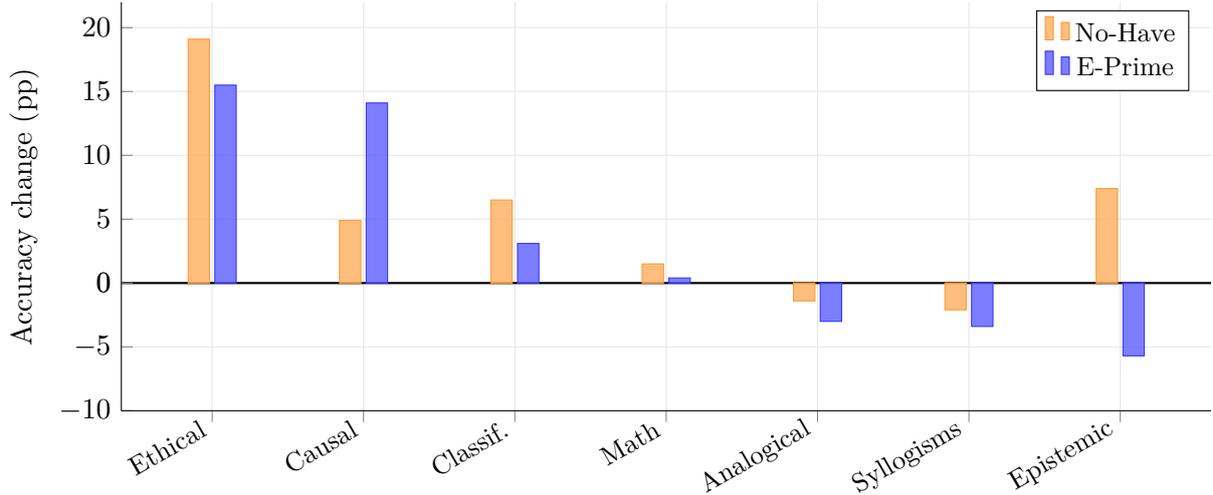
\begin{figure}[t]
\centering
\begin{tikzpicture}
\begin{axis}[
    ybar,
    bar width=8pt,
    width=\textwidth,
    height=7cm,
    ylabel={Accuracy change (pp)},
    symbolic x coords={Ethical,Causal,Classif.,Math,Analogical,Syllogisms,Epistemic},
    xtick=data,
    x tick label style={rotate=30, anchor=east, font=\small},
    ymin=-10, ymax=22,
    ytick={-10,-5,0,5,10,15,20},
    legend style={at={(0.98,0.98)}, anchor=north east, font=\small},
    grid=major,
    grid style={gray!20},
    every axis plot/.append style={fill opacity=0.85},
    nodes near coords={\empty},
    axis lines*=left,
    extra y ticks={0},
    extra y tick style={grid style={black, thick}},
]
\addplot[fill=orange!60, draw=orange!80] coordinates {
    (Ethical,19.1) (Causal,4.9) (Classif.,6.5) (Math,1.5)
    (Analogical,-1.4) (Syllogisms,-2.1) (Epistemic,7.4)
};
\addplot[fill=blue!60, draw=blue!80] coordinates {
    (Ethical,15.5) (Causal,14.1) (Classif.,3.1) (Math,0.4)
    (Analogical,-3.0) (Syllogisms,-3.4) (Epistemic,-5.7)
};
\legend{No-Have, E-Prime}
\end{axis}
\end{tikzpicture}
\caption{Crossover pattern of constraint effects across seven reasoning tasks (all models pooled). No-Have (orange) shows a more uniformly positive profile. E-Prime (blue) improves causal and ethical reasoning while degrading syllogisms and epistemic calibration. Tasks ordered by No-Have effect size.}
\label{fig:crossover}
\end{figure}

\subsubsection{Model-Specific Effects}

\begin{table}[t]
\centering
\caption{Selected model-specific constraint effects with gap-normalized percentages (Gap\% = delta / available improvement room). ``---'' indicates degradation or near-ceiling baseline where gap normalization is not meaningful. Full per-model breakdown in Appendix~D.}
\label{tab:model-effects}
\smallskip
\small
\begin{tabular}{@{}llrrrrc@{}}
\toprule
Model & Task & Ctrl & $\Delta$(NH) & Gap\%(NH) & $\Delta$(EP) & Gap\%(EP) \\
\midrule
Gemini Flash Lite & Ethical dilem.     & 41.7\% & +46.3 pp & 79.4\% & \textbf{+42.3 pp} & 72.4\% \\
Gemini Flash Lite & Causal reason.     & 57.8\% & +20.7 pp & 49.1\% & \textbf{+37.5 pp} & 88.7\% \\
Gemini Flash Lite & Epist.\ calib.     & 64.9\% & +22.2 pp & 63.4\% & +9.1 pp            & 25.9\% \\
Gemini Flash Lite & Classification     & 80.0\% & +18.8 pp & 93.8\% & +13.4 pp           & 67.0\% \\
GPT-4o-mini       & Ethical dilem.     & 91.7\% & +8.3 pp  & 100.0\% & +5.0 pp           & 60.0\% \\
GPT-4o-mini       & Causal reason.     & 75.6\% & $-$5.6 pp & ---   & +8.9 pp            & 36.5\% \\
GPT-4o-mini       & Epist.\ calib.     & 53.8\% & $-$3.7 pp & ---   & \textbf{$-$27.5 pp} & --- \\
Haiku 4.5         & Epist.\ calib.     & 89.0\% & +4.0 pp  & 36.8\% & +2.7 pp            & 24.5\% \\
Haiku 4.5         & All other tasks    & 83.8--100\% & $-$1.2 to +2.5 & --- & $-$4.7 to +1.2 & --- \\
\bottomrule
\end{tabular}
\end{table}

Two patterns demand attention. First, \textbf{No-Have is broadly beneficial across models}. Gemini shows large improvements (ethical +46.3~pp, epistemic +22.2~pp, classification +18.8~pp), GPT-4o-mini improves on ethical dilemmas (+8.3~pp), and even Haiku---which shows near-ceiling baselines and little room for improvement---shows small positive effects on epistemic calibration (+4.0~pp). No-Have's profile is consistent: it helps where possessive framing distorts, and it rarely hurts.

Second, \textbf{E-Prime is model-dependent in ways No-Have is not}. Gemini shows massive improvements under E-Prime (ethical +42.3~pp, causal +37.5~pp), but Gemini's low control baselines (41.7\% on ethical dilemmas, 57.8\% on causal reasoning---barely above chance on 4-option multiple-choice) mean the gap-normalized effects, while still large (72--89\% of available improvement), must be interpreted cautiously. GPT-4o-mini shows a mixed profile: E-Prime helps on causal reasoning (+8.9~pp) but \emph{devastates} epistemic calibration ($-27.5$~pp). Haiku shows small, mostly negative effects.

The GPT-4o-mini epistemic result is particularly informative. E-Prime eliminates the copula that structures calibrated hedging (``this claim is well-supported,'' ``the evidence is inconclusive''). For epistemic calibration tasks that require precisely this kind of graduated assertion, removing ``to be'' destroys the model's primary tool for expressing certainty levels---but only for GPT-4o-mini, suggesting this model relies more heavily on copula-based epistemic constructions than the others. No-Have, by contrast, produces only a mild $-3.7$~pp effect on the same task for the same model---evidence that possessive removal is a less disruptive intervention than copula removal.

Cross-model correlation of the E-Prime effect pattern (per-task accuracy delta): Gemini vs.\ GPT-4o-mini $r = 0.43$, Haiku vs.\ Gemini $r = -0.36$, Haiku vs.\ GPT-4o-mini $r = -0.75$ ($p \approx 0.05$, $n = 7$ tasks; suggestive but low-powered). The constraint does not produce a universal effect---it interacts with the model's training and architecture. Different models occupy different native Umwelten, and an imposed constraint alters each native Umwelt differently. The contrast with No-Have---which shows broadly positive effects regardless of model---suggests that possessive framing may represent a more universal cognitive default than copula-based identity assertion.

\subsubsection{Conciseness Effect}

\begin{table}[H]
\centering
\caption{Average word count by condition. Constraints reduce verbosity by 16--33\% across all non-mathematical tasks.}
\label{tab:wordcount}
\smallskip
\begin{tabular}{@{}lrrrr@{}}
\toprule
Task & Control & E-Prime & $\Delta$\% & No-Have \\
\midrule
Classification        & 407 & 273 & $-$33\% & 262 \\
Causal reasoning      & 570 & 396 & $-$31\% & 384 \\
Ethical dilemmas      & 557 & 437 & $-$22\% & 420 \\
Syllogisms            & 308 & 240 & $-$22\% & 227 \\
Epistemic calibration & 541 & 439 & $-$19\% & 416 \\
Analogical reasoning  & 366 & 308 & $-$16\% & 303 \\
Math word problems    & 182 & 178 & $-$2\%  & 178 \\
\bottomrule
\end{tabular}
\end{table}

The conciseness effect is the most robust finding across all three models and all seven tasks. Unlike accuracy, which varies by model and task, word count reduction under constraints is universal. Math word problems---which require numerical manipulation rather than verbal elaboration---show negligible reduction, confirming that the effect targets verbosity rather than essential reasoning content.

\subsubsection{Compliance}

No-Have violations occurred in 7.2\% of No-Have trials (111 of 1,538), with violations concentrated in low counts (1--2 per trial when present). E-Prime violations occurred in 51.9\% of E-Prime trials, with a mean of 1.7 violations per trial. This asymmetry is itself informative: ``to be'' pervades English far more deeply than ``to have,'' making E-Prime a fundamentally harder constraint to maintain. No-Have's 92.8\% compliance rate means its effects can be attributed to the constraint itself with substantially less interpretive ambiguity than E-Prime's.

Compliance-filtered analysis strengthens the pattern for both constraints. For E-Prime (zero violations only, $N{=}675$): the causal reasoning improvement remains strong (+13.1~pp vs.\ +14.1~pp unfiltered), ethical dilemmas strengthens (+16.9~pp vs.\ +15.5~pp), and math word problems shifts from neutral to +4.1~pp. The epistemic calibration degradation largely disappears ($-1.7$~pp vs.\ $-5.7$~pp), suggesting that non-compliant E-Prime trials---where the model struggles with constraint adherence---drive much of the epistemic degradation. For No-Have (zero violations only, $N{=}1{,}427$): the effect profile remains stable, consistent with its high baseline compliance. Full compliance amplifies the beneficial effects and attenuates the harmful ones under both constraints.

\textbf{Temperature note.} Experiment~1 used temperature~0.7 for three of four repetitions per item (one deterministic at 0.0, three stochastic at 0.7). This introduces sampling variance into per-condition accuracy estimates. The large sample size ($N{=}4{,}344$ scoreable trials) mitigates this, and the statistical tests account for the resulting variation. Experiment~2 (Section~4) used temperature~0.0 throughout, making its results deterministic.

\subsection{Limitations of Experiment 1}
\label{sec:exp1-limitations}

\textbf{Answer extraction.} Constrained responses sometimes use non-standard answer formats (``the strongest argument resides in Option~B,'' ``the answer lies in~B'') that require a more flexible extraction pipeline than standard regex patterns. An initial extraction pass missed 452 of 4,429 trials; after expanding the extractor to handle ``Option X'' format, relational phrasing, and \LaTeX{} boxed answers, extraction failures dropped to 85 (1.9\% of trials), distributed without strong systematic bias across conditions. The remaining failures are concentrated in ethical dilemmas and epistemic calibration for Haiku, where some responses embed the answer in discursive prose without any extractable marker.

\textbf{Ceiling and floor effects.} Syllogisms hit a ceiling at 100\% control accuracy (all models), compressing the observable degradation range. Gemini's control accuracy on causal reasoning (56.8\%) leaves room for large improvements that Haiku's 97.7\% baseline does not. Model-specific effects are partially confounded with baseline difficulty.

\textbf{Constraint semantics vs.\ constraint difficulty.} E-Prime's 51.9\% violation rate means observed effects reflect a mixture of compliant and non-compliant reasoning. That No-Have achieves 92.8\% compliance with stronger accuracy effects suggests that compliance difficulty and cognitive restructuring are at least partially independent dimensions.

\section{Experiment 2: Linguistic Orthogonality in Agent Ensembles}

\subsection{Motivation}

If linguistic constraints define different cognitive Umwelten, then agents operating under different constraints should perceive different features of the same problem---even when no individual constrained agent outperforms the control. The Umwelt framework predicts that cognitive diversity, operationalized through linguistic diversity, produces complementary coverage rather than redundant agreement.

\subsection{Design}

\textbf{Agents.} 16 agents, each defined by a system prompt encoding a single linguistic constraint. The control agent receives a standard debugging instruction with no linguistic constraint. Table~\ref{tab:agents} lists all agents with their constraint axes.

\begin{table}[t]
\centering
\caption{The 16 linguistic agents and their constraint axes.}
\label{tab:agents}
\smallskip
\small
\begin{tabular}{@{}lll@{}}
\toprule
Agent & Axis & Constraint Summary \\
\midrule
control           & ---                  & Standard English debugging \\
e\_prime          & ontological          & No ``to be'' verbs \\
quantified        & epistemic confid.    & All claims require confidence level \\
socratic          & reasoning struct.    & Reason through question-and-answer \\
steel\_man        & adversarial pos.     & State strongest case before critiquing \\
evidential        & epistemic source     & Tag each statement with derivation \\
temporal          & sequencing           & Step-by-step execution order \\
negation\_free    & specification        & No negation operators \\
constraint\_based & relational           & Express findings as constraints \\
analogical        & cross-domain         & Non-software analogy first \\
devils\_advocate  & adversarial neg.     & Assume bugs exist; construct failures \\
first\_principles & foundational         & Derive from axioms only \\
phenomenological  & perspectival         & Reason from data's first-person view \\
counterfactual    & modal                & State what would differ if false \\
minimal           & expressive range     & Fewest possible words \\
diachronic        & temporal-evol.       & Logic that outlived its context \\
\bottomrule
\end{tabular}
\end{table}

\textbf{Problems.} 17 software debugging problems across four categories: planted bugs~(6), logic puzzles~(2), specification ambiguities~(2), root cause analysis~(2), and miscellaneous~(5). Each problem includes ground-truth findings.

\textbf{Pipeline.} Five phases executed sequentially: (1)~parallel agent execution via \texttt{anthropic.AsyncAnthropic} at concurrency~8; (2)~LLM-based claim extraction from raw outputs; (3)~ground-truth matching via LLM judge (semantic, not string-exact); (4)~divergence map construction (convergent and unique clusters); (5)~orthogonality analysis (accuracy, Shapley values, pairwise redundancy, minimal ensemble selection).

\textbf{Model.} Claude Haiku~4.5 (\texttt{claude-haiku-4-5-20251001}) for all phases.

\subsection{Results}

\subsubsection{Individual Agent Accuracy}

\begin{table}[t]
\centering
\caption{Per-agent accuracy on 51 ground-truth findings across 17 problems. No constrained agent exceeds the control.}
\label{tab:agent-accuracy}
\smallskip
\small
\begin{tabular}{@{}p{24em}rr@{}}
\toprule
Agent(s) & Accuracy & $\Delta$ vs.\ Ctrl \\
\midrule
analogical, constraint\_based, control, counterfactual, phenomenological, steel\_man & 88.2\% & 0.0 \\
devils\_advocate, diachronic, temporal & 86.3\% & $-$1.9 \\
evidential, minimal & 84.3\% & $-$3.9 \\
first\_principles, negation\_free & 82.4\% & $-$5.8 \\
e\_prime, socratic & 80.4\% & $-$7.8 \\
quantified & 76.5\% & $-$11.7 \\
\bottomrule
\end{tabular}
\end{table}

\subsubsection{Ensemble Performance}

\begin{table}[H]
\centering
\caption{Ensemble accuracy. The full 16-agent ensemble achieves perfect ground-truth coverage. A 3-agent subset matches this ceiling.}
\label{tab:ensemble}
\smallskip
\begin{tabular}{@{}lr@{}}
\toprule
Configuration & Accuracy \\
\midrule
Control (single agent) & 88.2\% \\
Full ensemble (16 agents, union) & 100.0\% \\
Minimal ensemble (3 agents, greedy) & 100.0\% \\
\bottomrule
\end{tabular}
\end{table}

The minimal ensemble, selected by greedy Shapley-weighted addition, consists of: (1)~\textbf{analogical} (88.2\% individual, 8.5\% Shapley), (2)~\textbf{counterfactual} (88.2\% individual, 10.8\% Shapley), and (3)~\textbf{minimal} (84.3\% individual, 8.0\% Shapley). This 3-agent ensemble requires 17.6\% of the API calls of the full ensemble while achieving identical accuracy. All agents ran at temperature~0.0, making results deterministic---running the same ensemble again produces identical coverage.

\subsubsection{Permutation Test}
\label{sec:permutation}

To test whether the ensemble result depends on principled linguistic diversity or would emerge from any 3-agent subset, I evaluated all $\binom{16}{3} = 560$ possible 3-agent combinations. Of these, 45 (8.0\%) achieve 100\% coverage. The median 3-agent subset covers 96.1\% (49/51 findings), and the minimum covers 88.2\% (matching the best individual agent). The greedy-selected ensemble is one of 45 valid configurations, not a lucky outlier---but 92\% of random 3-agent subsets fail to achieve full coverage. Linguistic diversity matters, though it is not uniquely instantiated by the selected trio.

A structural constraint emerges: \textbf{every 100\%-coverage subset contains the counterfactual agent.} This is because counterfactual is the only agent that surfaced the hardest finding (specification ambiguity between first-occurrence and last-occurrence semantics---see Section~\ref{sec:unique}). Without counterfactual, 100\% coverage is impossible regardless of which other agents are included. The second most common member of perfect subsets is devils\_advocate (15/45), because the second-hardest finding (lost exception context in error wrapper) is covered by only three agents.

\subsubsection{Unique Contributions}
\label{sec:unique}

Only one agent contributed a finding that no other agent surfaced: \textbf{counterfactual} identified that a specification's use of ``preserving order'' was ambiguous between first-occurrence and last-occurrence semantics. This finding was surfaced by the counterfactual agent's explicit exploration of ``what would differ if this assumption were false''---a cognitive operation that the control agent's Umwelt did not afford. The permutation test (Section~\ref{sec:permutation}) confirms the structural importance of this unique contribution: without counterfactual, no 3-agent subset achieves 100\% coverage.

\subsubsection{Pairwise Redundancy}

Jaccard overlap between agent pairs ranged from 0.76 (control--counterfactual) to 0.93 (control--diachronic). Agents on similar constraint axes (temporal, diachronic, minimal) clustered at high overlap ($\geq 0.93$). Agents on distant axes (devils\_advocate--socratic, evidential--first\_principles) achieved lowest overlap (0.77), consistent with greater cognitive orthogonality.

\subsubsection{Shapley Values}

Shapley values ranged from 0.0703 (quantified, 7.0\%) to 0.1084 (counterfactual, 10.8\%), with 14 of 16 agents clustering between 7\% and 9\%. The relatively even distribution suggests that ensemble gain arises from broad complementarity rather than a single high-value agent---though counterfactual's unique contribution gives it an outsized share.

\subsection{Cross-Experiment Consistency}

E-Prime's effect differs between the two experiments. In Experiment~2 (software debugging, Haiku only), E-Prime degrades individual accuracy by $-7.8$~pp, consistent with Experiment~1's Haiku-specific pattern. In Experiment~1's multi-model aggregate, E-Prime improves ethical dilemmas (+15.5~pp) and causal reasoning (+14.1~pp) while degrading syllogisms ($-3.4$~pp). The divergence is informative: the debugging task's open-ended response format may interact differently with E-Prime than multiple-choice items, and the effect is model-dependent (Gemini drives most of the Experiment~1 improvements). The ensemble gain in Experiment~2---where linguistically diverse agents achieve 100\% coverage vs.\ 88.2\% for the control---demonstrates a mechanism (cognitive diversification) that operates independently of any individual constraint's accuracy effect.

\section{Discussion}

\subsection{Constraints Redirect, Not Merely Degrade}

The initial pilot ($N{=}70$, single model) suggested that linguistic constraints impose a uniform cognitive tax. The full experiment ($N{=}4{,}470$, three models, seven tasks, two constraints) reveals a more complex picture: constraints redirect reasoning in ways that are task-dependent, model-dependent, and constraint-dependent.

\textbf{No-Have's broad effectiveness.} The most practically significant finding is No-Have's consistent positive effect across tasks and models. Removing possessive framing improved 5 of 7 tasks, with the two largest effects---ethical dilemmas (+19.1~pp) and epistemic calibration (+7.4~pp)---on tasks where possession metaphors are densest (``the patient has a right,'' ``this claim has strong support''). No-Have achieved this while maintaining 92.8\% compliance, meaning its effects can be attributed to the constraint itself with minimal interpretive ambiguity. The crossover pattern is real but muted: only analogical reasoning ($-1.4$~pp) and syllogisms ($-2.1$~pp) showed declines, both non-significant.

\textbf{E-Prime's model-dependent volatility.} E-Prime shows the more dramatic effects, but they are unstable across models. The same constraint reshapes cognition in opposite directions depending on the model (Table~\ref{tab:model-effects}), with cross-model correlations reaching $r = -0.75$ ($p \approx 0.05$, $n = 7$; suggestive). This only makes sense if different models occupy different \emph{native Umwelten}---default cognitive worlds established by training corpus, architecture, and alignment process. An imposed constraint interacts with this native world rather than overriding it. The crossover pattern---improvements on causal reasoning and ethical dilemmas, degradation on syllogisms and epistemic calibration---is a signature that the constraint is redirecting cognition rather than generically improving or impairing it, though the absence of an active control matching the constraint prompt's elaborateness means a metalinguistic self-monitoring confound cannot be fully excluded (see Section~\ref{sec:limitations}).

\textbf{The constraint comparison.} No-Have outperforms E-Prime on 5 of 7 tasks despite being theoretically less studied and receiving no predictions of superiority. One interpretation: ``have'' encodes ownership metaphors that are particularly distorting for abstract reasoning (a patient ``has'' rights, an argument ``has'' flaws), while ``to be'' is so pervasive that its elimination creates more noise than signal. Another: No-Have's higher compliance (92.8\% vs.\ 48.1\%) means the restructuring operates more cleanly. The compliance-filtered analysis supports this: fully compliant E-Prime trials show stronger benefits and attenuated harms, suggesting the constraint's theoretical mechanism works but is diluted by compliance failures.

The ensemble results from Experiment~2 provide a complementary mechanism. No individual constrained agent exceeds the control's 88.2\% accuracy on software debugging, yet a union ensemble achieves 100\% coverage and a 3-agent subset matches this ceiling. The constraints produce \emph{orthogonal coverage patterns}---different agents perceive different features of the same problem. This diversification effect operates independently of any individual constraint's accuracy impact.

Together, the two experiments demonstrate that linguistic constraints operate through at least two mechanisms: (1)~\emph{cognitive restructuring}, where removing a linguistic default forces more explicit reasoning---No-Have demonstrates this most cleanly, with broad improvements and high compliance---and (2)~\emph{cognitive diversification}, where different constraints activate different regions of the model's latent reasoning capacity (Experiment~2: ensemble coverage). These mechanisms are not mutually exclusive---a constraint can restructure reasoning for one task while providing orthogonal coverage when combined with other constraints.

The stronger claim from the pilot---that constraints are merely a cognitive tax---does not survive multi-model testing. But neither does the naive expectation that constraints uniformly help. The Umwelt is not a dial that turns reasoning up or down; it is a lens that brings some features into focus and blurs others.

\subsection{Native Umwelten and Model-Constraint Interaction}

The model-dependent pattern demands interpretation. Why does Gemini benefit from E-Prime while Haiku does not?

One hypothesis: models differ in how tightly their default reasoning is coupled to copula-based formulations. If Gemini's training over-relies on ``X is Y'' patterns for classification and causal attribution, then E-Prime forces a beneficial decoupling. If Haiku's training has already diversified away from copula-dependence---perhaps through Anthropic's RLHF process or constitutional AI training---then E-Prime removes useful structure without providing compensatory restructuring.

A related hypothesis: Gemini's lower baseline accuracy on several tasks (causal reasoning: 57.8\% vs.\ Haiku's 97.7\%) leaves more room for improvement. The constraint may function as a form of implicit chain-of-thought: by forcing circumlocution, it increases the reasoning steps between stimulus and answer, benefiting models that would otherwise shortcut to an incorrect response. Models that already reason carefully gain less from this forced elaboration.

The GPT-4o-mini epistemic result provides a third data point. E-Prime collapses GPT-4o-mini's epistemic calibration from 53.8\% to 26.2\%---a $-27.5$~pp effect that dwarfs any other single-model degradation in the experiment. This suggests GPT-4o-mini relies heavily on copula constructions (``this claim is well-supported,'' ``the evidence is inconclusive'') as its primary mechanism for graduated epistemic assertion. Removing these constructions doesn't merely add friction---it eliminates the model's epistemic vocabulary. Haiku, by contrast, shows a +2.7~pp \emph{improvement} on the same task under E-Prime, suggesting it has alternative epistemic strategies that activate when the copula pathway is blocked.

These hypotheses are testable. Attention analysis on constrained vs.\ unconstrained inference would reveal whether E-Prime activates different internal circuits (the Umwelt interpretation) or merely adds output-level variance (the cognitive tax interpretation). The GPT-4o-mini epistemic collapse suggests the former: a model that has \emph{no} alternative pathway for a cognitive operation will fail catastrophically when its primary pathway is blocked, rather than degrading gradually as a noise account would predict. This distinction---activation of latent strategies vs.\ statistical noise---is perhaps the most important open question the Umwelt framework raises.

\subsection{Why Possessive Framing Distorts More Than Identity Framing}

No-Have was originally included as an exploratory control: a second constraint to test whether E-Prime's effects were specific to copula elimination or generalized to any vocabulary restriction. The results invert this framing---No-Have is the more effective and more consistent intervention.

No-Have's +19.1~pp improvement on ethical dilemmas ($d = 0.57$, $p < 0.001$) is the largest aggregate effect in the experiment. The mechanism is plausible: ethical reasoning in natural language is saturated with possessive framing---patients ``have'' rights, actions ``have'' consequences, stakeholders ``have'' interests. This framing reifies abstract relationships as owned properties, potentially obscuring the relational structure that ethical analysis requires. Removing ``have'' forces the model to articulate these relationships explicitly: ``this action affects the patient's autonomy'' rather than ``the patient has a right.'' The same mechanism plausibly explains the classification improvement (+6.5~pp): categories ``have'' members, objects ``have'' properties---possessive framing collapses relational structure into containment metaphors.

No-Have's 92.8\% compliance rate eliminates much of the interpretive ambiguity that plagues E-Prime analysis. When 51.9\% of E-Prime trials contain violations, observed effects reflect a messy mixture of compliant and non-compliant reasoning. No-Have's cleaner compliance means its effects can be attributed to the constraint itself rather than to partial compliance artifacts.

The practical implication is direct: for most reasoning tasks, No-Have is a more effective cognitive intervention than E-Prime---broader in its benefits, milder in its degradations, and far easier for models to maintain. E-Prime remains the more theoretically informative constraint, precisely because its volatile, model-dependent effects reveal the structure of native Umwelten. But as a tool for improving agent reasoning, No-Have is the stronger instrument.

\subsection{The Cognitive Tax Revisited}

The pilot suggested a uniform cognitive tax of ${\sim}8.7$~pp. The full experiment complicates this picture. E-Prime imposes a tax on syllogisms ($-3.4$~pp), epistemic calibration ($-5.7$~pp), and analogical reasoning ($-3.0$~pp), but produces gains on causal reasoning (+14.1~pp) and ethical dilemmas (+15.5~pp) that exceed any plausible tax. The net effect depends on the task.

A revised account: constraints impose two opposing forces. First, a \emph{compliance cost}---the computational overhead of monitoring and reformulating language, which degrades performance on all tasks. Second, a \emph{restructuring benefit}---the forced reformulation activates more explicit or more careful reasoning, which benefits tasks where default language masks reasoning gaps. The observed effect is the sum. For syllogisms, where the default language aligns well with the task structure, the compliance cost dominates. For causal reasoning and ethical dilemmas, where default language enables superficial pattern-matching, the restructuring benefit dominates.

The conciseness effect (16--33\% word reduction) is consistent with this account. Constraints eliminate filler and hedging, producing more efficient reasoning chains. The compression is universal across tasks and models, suggesting it reflects compliance cost (less capacity for elaboration) and restructuring benefit (less need for elaboration when reasoning is more focused) in combination.

\subsection{Counterfactual as Cognitive Affordance}

The counterfactual agent's unique finding---identifying specification ambiguity by asking ``what would differ if this assumption were false''---is a direct demonstration of a linguistic affordance creating a cognitive capability. The control agent had access to the same information and presumably knows what counterfactual reasoning is---it was not \emph{incapable} of the operation. But its Umwelt did not make that operation a default mode of perception. The constraint made systematic assumption-inversion the agent's habitual lens, and a finding followed that no other lens surfaced.

This distinction matters for the prompt-vs-Umwelt boundary discussed in Section~\ref{sec:frameworks}. One could argue that ``consider counterfactuals'' is simply a task instruction. But the counterfactual agent was not told to look for specification ambiguity---it was told to reason counterfactually about everything. The specific finding emerged because the cognitive mode made a specific feature of the problem perceptible. The constraint structured perception; the finding was a consequence.

\subsection{Implications for Agent Architecture}

If cognitive diversity is the mechanism underlying ensemble gain, then agent ensemble design becomes a question of Umwelt selection: which set of linguistic constraints produces maximally orthogonal coverage for a given task domain? The greedy selection algorithm identified analogical, counterfactual, and minimal as the optimal 3-agent subset for software debugging---three constraints drawn from three different axes (cross-domain mapping, modal reasoning, and expressive compression). This suggests that axis diversity, not constraint intensity, drives ensemble value.

The practical implication is that multi-agent systems should be designed not by duplicating capable agents, but by equipping agents with linguistically diverse reasoning modes. Three agents with different Umwelten outperform sixteen agents with overlapping ones.

\subsection{The Constraint Design Space}

These experiments tested a handful of constraints drawn from a much larger design space. Section~2.6 surveyed eight intellectual traditions, each of which identified a specific axis along which language shapes thought and proposed a linguistic intervention. These traditions were developed independently---Korzybski working on map-territory confusion, Bohm on process metaphysics, Bridgman on operationalization, Elgin on epistemic transparency, N\={a}g\={a}rjuna on non-binary logic---yet they converge on a shared structural insight: that linguistic defaults encode cognitive defaults, and that reforming the language reforms the cognition.

Table~\ref{tab:taxonomy} organizes these traditions as a constraint taxonomy for Umwelt engineering.

\begin{table}[t]
\centering
\caption{Taxonomy of linguistic constraints for Umwelt engineering, organized by intellectual tradition and targeted cognitive failure mode.}
\label{tab:taxonomy}
\smallskip
\small
\begin{tabular}{@{}llll@{}}
\toprule
Constraint & Tradition & Axis & Targets \\
\midrule
E-Prime            & Korzybski/Bourland & Semantic          & False identity claims \\
Gen.\ Semantics    & Korzybski          & Extensional       & Over-generalization \\
Rheomode           & Bohm               & Ontological       & Entity bias \\
Operationalism     & Bridgman           & Epist.-procedural & Ungrounded claims \\
Toki Pona          & Lang               & Lexical           & Abstraction leakage \\
Evidentiality      & Elgin/L\'{a}adan   & Epist.-source     & Unsourced confidence \\
Catu\d{s}ko\d{t}i & N\={a}g\={a}rjuna  & Logical           & Premature binary resol. \\
NVC                & Rosenberg          & Evaluative        & Obs.--judgment conflation \\
\bottomrule
\end{tabular}
\end{table}

Several features of this taxonomy bear emphasis. First, the axes are largely independent: removing identity claims (E-Prime) says nothing about evidential sourcing (L\'{a}adan), which says nothing about binary logic (Catu\d{s}ko\d{t}i). This independence predicts that constraints drawn from different axes will produce orthogonal effects on reasoning---precisely the mechanism that drove ensemble gain in Experiment~2. Second, each constraint makes a specific, testable prediction about which tasks it will improve and which it will degrade. E-Prime should degrade tasks that depend on identity bridges (confirmed: syllogisms) and improve tasks where identity claims mask reasoning gaps. Evidentiality constraints should improve tasks where epistemic sourcing matters (research synthesis, factual claims) and impose overhead on tasks where all information comes from a single authoritative source. The catu\d{s}ko\d{t}i should improve ethical dilemmas and design tradeoffs where binary framing loses information, and add unnecessary complexity to tasks with genuinely binary answers.

Third, the traditions suggest that the design space is not arbitrary. Each constraint was developed by careful thinkers who identified a real cognitive failure mode and proposed a linguistic remedy. The constraints have theoretical motivation, not just empirical novelty. This distinguishes Umwelt engineering from unprincipled prompt variation: the question is not ``what random linguistic constraints produce interesting effects?''\ but ``which established theories of language-thought interaction yield productive cognitive interventions for artificial agents?''

\subsection{Relationship to Existing Frameworks}
\label{sec:frameworks}

An obvious objection: if the constraint is delivered as a system prompt instruction, how is this not simply prompt engineering? The distinction requires careful articulation.

A prompt instruction specifies a \emph{task} within the agent's existing cognitive world. ``Reason step by step'' triggers a reasoning strategy; ``be concise'' adjusts an output parameter; ``you are a careful logician'' activates a behavioral persona. In each case, the conceptual vocabulary remains standard English---the agent applies the instruction using its full default repertoire of concepts and grammatical structures. An Umwelt intervention restructures the \emph{medium} through which all tasks are processed. ``Eliminate all forms of `to be'\,'' does not specify what to think about or how carefully to think---it removes an entire class of cognitive operations (identity assertion, categorical attribution, essentialist shorthand) from the agent's available repertoire, forcing all subsequent reasoning through alternative pathways.

The empirical evidence supports this distinction on four grounds. First, the model-dependent effects. ``Reason step by step'' does not produce negative cross-model correlations---it helps broadly, because it is a task-level instruction that interacts minimally with model-specific internal structure. E-Prime produces correlation coefficients of $r = -0.36$ and $r = -0.75$ between model pairs, meaning the same constraint reshapes cognition in opposite directions depending on the model's native architecture. This interaction signature is consistent with an intervention that engages internal representational structure rather than merely adding an output-level directive. Second, the GPT-4o-mini epistemic collapse ($-27.5$~pp on a single task) is not the gradual degradation that instruction-following overhead would produce---it is a catastrophic failure of a specific cognitive capacity, consistent with the removal of a load-bearing linguistic structure rather than the addition of a processing burden. Third, the compliance-filtered analysis shows that \emph{fully compliant} E-Prime trials produce stronger beneficial effects and attenuated harmful effects compared to unfiltered trials. If the constraint operated merely as an instruction competing for the model's attention, higher compliance would mean higher attentional cost and worse performance uniformly. Instead, higher compliance amplifies the restructuring benefit---the constraint is not taxing the reasoning; it is redirecting it.

Fourth, and most directly: \textbf{the two constraints function as mutual active controls for prompt elaborateness.} Both E-Prime and No-Have prompts are comparably elaborate---both list forbidden forms, provide reformulation examples, and impose a metalinguistic self-monitoring demand. If the observed effects arose from the general demand for self-monitoring rather than from the specific vocabulary restriction, the two constraints should produce similar task profiles. They do not. On epistemic calibration, No-Have improves accuracy by 7.4~pp while E-Prime degrades it by 5.7~pp---a 13.1~pp swing between two equally elaborate prompts. On causal reasoning, E-Prime improves by 14.1~pp while No-Have improves by only 4.9~pp. On classification, No-Have gains 6.5~pp to E-Prime's 3.1~pp. These differential effects can only be explained by \emph{which words} are being restricted---possessive framing versus copula-based identity assertion---not by the shared demand for linguistic self-monitoring. The within-study comparison controls for prompt elaborateness more directly than any external active control could, because the two conditions share every feature except the specific vocabulary targeted.

None of this proves that the three-layer distinction is ontologically real rather than a useful abstraction. But the empirical signatures---model-specific interaction, catastrophic capacity failure, compliance-benefit correlation, and divergent task profiles between equally elaborate constraints---are more consistent with a medium-level intervention than with a task-level instruction. The three-layer stack may ultimately reduce to a spectrum rather than a sharp hierarchy. Even so, the far end of that spectrum---where linguistic interventions interact with model internals in structured, model-dependent ways---represents a design space that prompt engineering as currently practiced does not address.

Returning to the three-layer stack introduced in Section~1.2: prompt engineering optimizes within a fixed Umwelt, context engineering provides information within a fixed Umwelt, and Umwelt engineering designs the Umwelt itself. The empirical signatures reported here---model-specific interaction patterns, catastrophic capacity failures, compliance-benefit correlations---populate this framework with evidence that the third layer is not merely conceptual.

The relationship to ORION's Mentalese \citep{tanmay2025} is direct: Mentalese is a deliberately designed Umwelt---a synthetic cognitive environment optimized for mathematical reasoning. The relationship to Coconut \citep{hao2024} is contrastive: Coconut demonstrates that language-based Umwelten may impose unnecessary constraints, suggesting that the design space includes non-linguistic cognitive environments. Both are instances of Umwelt engineering, whether their authors describe them as such or not.

\section{Limitations}
\label{sec:limitations}

\textbf{No external active control for prompt complexity (primary limitation).} The E-Prime and No-Have system prompts are substantially more elaborate than the control condition, which receives no constraint instruction. This introduces a confound: some portion of the observed effects could arise from the presence of an elaborate meta-cognitive instruction---one that demands self-monitoring of language output---rather than from the specific vocabulary restriction. The strongest defense is the within-study comparison: the two constraints are comparably elaborate but produce divergent task profiles (a 13.1~pp swing on epistemic calibration alone; see Section~\ref{sec:frameworks}), isolating the contribution of the specific vocabulary restriction from the shared self-monitoring demand. Nevertheless, a non-vocabulary active control (e.g., ``ensure every paragraph opens with a topic sentence and closes with a transition'' or ``keep all sentences under 15 words'') would provide external confirmation and rule out the possibility that any vocabulary-targeting prompt, regardless of which words are targeted, produces similar patterns. A dedicated active control experiment remains the most important next step for this research program.

\textbf{Ceiling and floor effects.} Syllogisms hit 100\% control accuracy for all three models, compressing the observable degradation range. Gemini's lower baseline on several tasks (causal reasoning: 57.8\%, ethical dilemmas: 41.7\%) inflates observable improvement relative to Haiku's higher baselines (97.7\%, 98.2\%). Table~\ref{tab:model-effects} reports gap-normalized effects alongside raw deltas to aid interpretation: Gemini's +42.3~pp ethical dilemmas improvement from a 41.7\% baseline represents 72.4\% of available improvement room, while Haiku's +2.7~pp from 89.0\% represents 24.5\%. Both are real effects, but the raw numbers are not directly comparable.

\textbf{E-Prime compliance.} The 51.9\% E-Prime violation rate means observed E-Prime effects reflect a mixture of compliant and non-compliant reasoning. Compliance-filtered analysis (Section~3.2.4) confirms the direction of effects but has reduced power. No-Have's 92.8\% compliance provides substantially cleaner causal evidence for constraint effects.

\textbf{Statistical notes.} $p$-values in Table~\ref{tab:aggregate} use Fisher's exact test (two-sided), appropriate for contingency tables with zero or small cells. The syllogisms results are affected by the 100\% control ceiling: E-Prime syllogisms degradation ($p = 0.015$) survives correction; No-Have syllogisms degradation ($p = 0.074$) does not reach conventional significance. Cohen's $d$ values are reported for comparability with the continuous-outcome literature but are approximate for binary data; odds ratios would be more conventional. Cross-model correlations ($r = -0.75$, $-0.36$, $0.43$) are computed on $n = 7$ tasks and should be interpreted as suggestive; the strongest ($r = -0.75$, Haiku vs.\ GPT-4o-mini) has $p \approx 0.05$.

\textbf{Model selection.} All three models are cost-efficient instruction-following models. Constraint effects may differ on frontier models (GPT-4o, Claude Sonnet/Opus, Gemini Pro), which may have more capacity for simultaneous constraint compliance and reasoning. The model-dependent effects observed here predict that frontier models will show different interaction patterns, not necessarily smaller effects.

\textbf{Task format.} All tasks use multiple-choice format for scoring consistency. Open-ended reasoning tasks---like Experiment~2's software debugging---may show different constraint effects. The conciseness finding (16--33\% word reduction) suggests that constraints alter reasoning structure, not just answer selection, but the multiple-choice format may undercount effects that manifest in reasoning quality rather than answer accuracy.

\textbf{LLM-as-judge.} Experiment~2 used an LLM judge for semantic matching of claims to ground truth. While this avoids brittle string matching, it introduces judge noise and potential biases.

\textbf{Answer extraction.} Despite iterative expansion of the extraction pipeline, 85 of 4,429 trials (1.9\%) resisted answer extraction. The remaining failures are concentrated in Haiku's ethical dilemma and epistemic calibration responses, where answers are embedded in discursive prose. While the overall impact is small, any systematic relationship between extraction failure and response correctness could bias accuracy estimates.

\section{Future Work}

The Umwelt framework opens several research directions:

\textbf{Constraint cartography.} The taxonomy in Table~\ref{tab:taxonomy} identifies eight constraints across seven axes; Experiment~1 tests two of these (E-Prime and No-Have) across seven task types, with results that confirm the existence of task-dependent crossover effects. A full cartography would cross all eight constraints with the same task battery, producing an $8 \times 7$ matrix of effects. Each cell encodes a testable prediction: evidentiality constraints should improve epistemic calibration tasks but impose overhead on single-source reasoning; the catu\d{s}ko\d{t}i should improve ethical dilemmas but add noise to tasks with binary ground truth; Toki Pona should improve explanation tasks but degrade tasks requiring precise technical vocabulary. The model-dependent effects in Experiment~1 add a third dimension: each cell may vary across model architectures, suggesting a constraint $\times$ task $\times$ model tensor rather than a simple matrix.

\textbf{Umwelt composition.} Can constraints be productively combined? An agent reasoning in E-Prime + required uncertainty markers + analogical framing operates in a more structured Umwelt than any single constraint provides. Whether constraints compose additively, interfere, or interact non-linearly is an empirical question.

\textbf{Dynamic Umwelt switching.} Should an agent's linguistic world change depending on what it is doing? Sketch-of-Thought's per-task routing \citep{sketch2025} suggests yes, but their paradigms are fixed. A richer version would allow agents to shift Umwelten mid-task as reasoning demands change.

\textbf{Native Umwelt characterization.} The model-dependent effects in Experiment~1 suggest that each model has a ``native Umwelt''---a default cognitive world established by its training corpus, architecture, and alignment process. Characterizing these native Umwelten is a prerequisite for principled constraint selection. If a model's native Umwelt already de-emphasizes copula-based reasoning (as Haiku's results suggest), E-Prime adds noise rather than restructuring. Mechanistic interpretability methods---sparse autoencoders, activation patching, probing classifiers---could map the native Umwelt of a model by identifying which linguistic patterns most strongly activate its reasoning circuits.

\textbf{Emergent vs.\ designed Umwelten.} Quiet-STaR and Coconut demonstrate that models can develop effective reasoning formats without human design. When should one impose a designed Umwelt, and when should one let it emerge? The trade-off between interpretability (designed) and optimality (emergent) is largely unexplored.

\textbf{Umwelt evaluation.} How do you measure whether one Umwelt is better than another for a given purpose? Accuracy alone is insufficient---the experiments here show that constrained agents may score lower individually while contributing more to ensemble coverage. Metrics for cognitive diversity, orthogonality of perception, and complementary coverage are needed.

\textbf{Cross-linguistic Umwelten.} If training language shapes LLM cognition \citep{wang2025}, then multilingual reasoning environments constitute natural Umwelt experiments. An agent that reasons in Japanese about a problem described in English operates in a different cognitive world than one reasoning entirely in English.

\textbf{Constraint space geometry.} The taxonomy in Table~\ref{tab:taxonomy} treats constraints as discrete categories, but the underlying space may be continuous---or at least partially so. Within a constraint family, one can titrate strictness (strict E-Prime vs.\ allowing copula in direct quotes vs.\ merely flagging identity claims). Across families, the question becomes whether constraints define positions in a shared geometric space with measurable axes and distances. If so, the orthogonality between constraints---whether E-Prime and evidentiality marking produce independent effect profiles across tasks, or whether their effects correlate---becomes empirically testable through factor analysis or representational similarity analysis on the constraint $\times$ task effect matrix. The ensemble results in Experiment~2 already suggest that constraints drawn from different axes produce more complementary coverage, but a systematic measurement of the constraint space's dimensionality and orthogonal structure would transform the taxonomy from a list into a coordinate system. This requires testing substantially more constraints (at minimum 6--8) across the same task battery---a natural sequel to the cartography program described above. A preliminary question is whether the space is genuinely continuous (admitting interpolation between constraints), mixed (continuous within families, discrete between them), or fundamentally discrete (with orthogonality measurable only as statistical independence of effect profiles). The answer determines whether Umwelt design is a search problem over a smooth manifold or a combinatorial problem over a structured set.

\section{Conclusion}

For a language model, the available language is not a transparent medium through which cognition passes---it \emph{is} the cognition. A human can think beneath and beyond their words; a standard LLM cannot. Its vocabulary determines which concepts exist. Its grammar determines which relationships between concepts are expressible. Its conceptual distinctions determine which features of a problem become perceptible. Designing this language is Umwelt engineering: the construction of cognitive worlds for artificial minds.

The experiments demonstrate that linguistic constraints reshape agent cognition in measurable, task-dependent, and model-dependent ways. Removing possessive ``to have''---a constraint that was originally exploratory---produces the broadest improvement: ethical dilemmas +19.1~pp, classification +6.5~pp, epistemic calibration +7.4~pp, with 92.8\% compliance and consistent effects across models. Removing ``to be'' produces more dramatic but less predictable effects: causal reasoning +14.1~pp and ethical dilemmas +15.5~pp, but model-dependent volatility so severe that cross-model correlations of E-Prime effects reach $r = -0.75$. The contrast between the two constraints is itself revealing: possessive framing appears to be a more universal cognitive default than copula-based identity assertion, producing a broader distortion that can be more cleanly removed. These effects replicate across three models from three vendors, though the magnitude and even direction vary by model---revealing that each model occupies a different native Umwelt that interacts differently with imposed constraints. In multi-agent settings, linguistically diverse agents achieve coverage that no individual agent can match: a 3-agent ensemble selected for Umwelt diversity achieves 100\% ground-truth coverage on a debugging task where the best individual agent reaches 88.2\%---and a permutation test confirms that only 8\% of random 3-agent subsets match this ceiling, with every successful subset containing the counterfactual agent.

Two mechanisms emerge. \emph{Cognitive restructuring}: constraints that remove linguistic defaults force more explicit, operational reasoning---No-Have demonstrates this most cleanly, with broad improvements and high compliance, while E-Prime reveals the mechanism's limits when compliance is low and model interaction is high. \emph{Cognitive diversification}: different constraints activate different regions of a model's latent reasoning capacity, producing orthogonal coverage in ensemble settings---demonstrated by the counterfactual agent's unique finding and confirmed by the permutation test. Both mechanisms confirm that the linguistic cognitive environment determines the space of possible thought---an empirically measurable design variable, not a philosophical speculation. The primary open question is whether the observed restructuring effects are driven by the specific vocabulary restrictions or by the general demand for metalinguistic self-monitoring that any elaborate constraint prompt imposes; the crossover pattern favors the former but cannot rule out the latter without an active control experiment.

These findings, together with converging evidence from synthetic reasoning languages \citep{tanmay2025}, latent-space reasoning \citep{hao2024}, and cross-linguistic cognition in LLMs \citep{wang2025}, establish the case for a three-layer framework---prompt engineering, context engineering, Umwelt engineering---and call for systematic investigation of the design space it opens.

\bigskip
\noindent\emph{Design the world first. Then worry about the question.}

\bibliography{references}

\appendix

\section{E-Prime Constraint Prompt}

The following system prompt was used for the E-Prime condition in Experiment~1:

\begin{quote}
You must reason and respond entirely in E-Prime---a form of English that eliminates all forms of ``to be.'' You may not use: is, am, are, was, were, be, being, been, or contractions containing these (it's, that's, there's, who's, etc.). Reformulate all statements using active verbs, process descriptions, or relational language. Do not merely rephrase surface syntax---restructure your reasoning to avoid ontological identity claims.
\end{quote}

\section{Agent Constraint Prompts}

Full system prompts for all 16 agents in Experiment~2 are available in the supplementary repository.

\section{No-Have Constraint Prompt}

The following system prompt was used for the No-Have condition in Experiment~1:

\begin{quote}
You must reason and respond without using any form of ``to have'' as a main verb. You may not use: has, have, had, having when they express possession, containment, or attribution. Auxiliary uses are permitted (e.g., ``has completed,'' ``have been''). Reformulate all possessive statements using relational, behavioral, or structural language. ``The argument has a flaw'' becomes ``a flaw appears in the argument.'' ``This system has three components'' becomes ``three components make up this system.''
\end{quote}

\section{Per-Model Accuracy Breakdown}

Full per-model accuracy tables are available in the supplementary data files.

\section{Reproducibility}

All code, data, and results are available at: \url{https://github.com/rodspeed/umwelt-engineering}.

\begin{itemize}[nosep]
  \item Experiment~1: \texttt{e-prime-llm/}---task items, scoring rubrics, multi-model runner, and full results
  \item Experiment~2: \texttt{linguistic-agents/}---agent definitions, problem bank, 5-phase pipeline, and full analysis
  \item Models: Claude Haiku~4.5 (\texttt{claude-haiku-4-5-20251001}), GPT-4o-mini (\texttt{gpt-4o-mini-2024-07-18}), Gemini~2.5 Flash Lite (\texttt{gemini-2.5-flash-lite})
  \item Experiment~1: 4,470 trials across 7 tasks $\times$ 3 conditions $\times$ 3 models $\times$ 4 repetitions
  \item All experiments are resumable (JSONL append) and reproducible at temperature~0.0
\end{itemize}

\section*{Acknowledgments}

This paper was developed through extensive collaboration with Claude (Anthropic, 2024--2026). The Umwelt engineering framework, three-layer stack, experimental design, and core thesis are the author's own. The constraint taxonomy (Table~\ref{tab:taxonomy}) emerged from directed inquiry: the author hypothesized that a design space of linguistic cognitive constraints existed beyond E-Prime and used structured dialogue with Claude to surface candidate traditions, which were then evaluated, organized, and integrated into the framework by the author. Claude also assisted with literature review, drafting, statistical analysis, and code for the experimental pipeline. This division of labor---human hypothesis and architectural judgment, AI recall and drafting---is itself an instance of the collaborative cognitive environments this paper examines.

\end{document}